\documentclass{article}
\usepackage{spconf,amsmath,graphicx,hyperref}

\usepackage{amsmath}
\usepackage{multirow}
\def\textem#1{{\em{#1}}}%

\title{Equipping Large Language Model with Directional Speech Understanding Capabilities}
\name{Ju Lin, Jing Pan, Ruizhi Li, Ming Sun, Yuzong Liu, Alaa Hassan, Jing Zheng, Florian Metze}
\address{Meta, USA}
\begin{document}
%
\maketitle
\begin{abstract}
Recent studies have demonstrated that prompting large language models (LLM) with audio encodings enables effective
speech understanding capabilities. However, most speech LLMs are trained on single-channel, single-talker data, which makes it challenging to directly apply them to multi-talker and multi-channel speech understanding task. In this work, we present a comprehensive investigation on how to enable directional multi-talker speech understanding capabilities for LLMs, specifically in smart glasses usecase. We propose two novel approaches to integrate directivity into LLMs: (1) a cascaded system that leverages a source separation front-end module, and (2) an end-to-end system that utilizes serialized output training. All of the approaches utilize a multi-microphone array embedded in smart glasses to optimize directivity interpretation and processing in a streaming manner. Experimental results demonstrate the efficacy of our proposed methods in endowing LLMs with directional speech understanding capabilities, achieving strong performance in both speech recognition and speech translation tasks.
\end{abstract}
\section{Introduction}
\label{sec:intro}

Automatic transcription of a conversation partner from several feet away is a crucial emerging scenario in Automatic Speech Recognition (ASR), particularly in multi-talker environments. Wearable devices, such as smart glasses, can leverage this technology to generate real-time captions, enabling a range of innovative applications, including speech translation, and enhancing communication for individuals with hearing impairments. Recent advances~\cite{lin2023directional, subramanian2021directional, feng2023directional, lin2025directional,lin2024agadir} in Directional Automatic Speech Recognition (DASR) have demonstrated the effectiveness of using multi-microphone arrays to identify and separate speakers, including the wearer, conversation partners, and background bystanders, while also exhibiting improved robustness to noise compared to traditional single-channel beamforming approaches.

More recently, the advent of speech language models (SLMs) has demonstrated tremendous potential in a wide range of speech-related tasks, including automatic speech recognition (ASR), speech translation, speaker diarization, and speech synthesis\cite{arora2025landscape, hu2024wavllm, gong2023listen}. However, a key challenge in these tasks is handling multi-talker speech with overlap. Serialized Output Training (SOT) method~\cite{kanda2020serialized} was proposed to address this challenge. SOT serializes reference transcripts according to speaker start times and inserts a special speaker change token between reference segments. Several recent studies have proposed innovative LLM-based SOT methods for multitalker ASR. For example, Meng et al.\cite{meng2025large} introduced an LLM-based SOT approach that leverages pre-trained speech encoders and LLMs, fine-tuning them on single-channel multi-talker datasets. Similarly, Xie et al.\cite{xie2025thinking} proposed Directional-SpeechLlama, which utilizes serialized directional output training to perform speech recognition and source localization tasks. Furthermore, Shi et al.~\cite{shi2025serialized} introduced Serialized Output Prompting (SOP), an adaptive prompting method that improves performance in overlapping cases by indicating how multi-talker contents are mixed and can be separated according to the input.

In this paper, we present two novel approaches to enhance SLMs with the capability to process and understand directional speech, focusing on smart glasses applications. Specifically, we consider scenarios where wearer (near-field source) engage in conversations with a distant conversation partner (far-field source), resulting in distinct spatial acoustic differences that can be exploited through multi-channel processing. We propose a two-stage approach, where we first employ a streaming multi-channel directional source separation model to separate the audio stream into individual speaker segments, and then apply post-processing to predict the speaker tag for each small audio chunk. The predicted speaker tag is subsequently used to select a task-specific prompt, such as "translate to target language" for translation tasks, thereby enabling the model to generate targeted responses tailored to the specific speaker and task at hand. In addition, we propose an SOT-based approach that involves fine-tuning LLMs to accept multichannel audio as input using SOT-style training data. This approach enables the model to incorporate directional information from beamformed signals, allowing it to follow instructions specific to individual speakers and facilitate personalized, speaker-dependent processing. Furthermore, for both approaches, we enable streaming inference for the SLM, supporting simultaneous speech recognition and translation.

\section{Proposed Approaches}
\label{sec:format}
\subsection{Streaming Directional Source Separation Model}
\begin{figure}[!h]
\vspace{-2mm}
    \centering
    \includegraphics[width=0.95\linewidth]{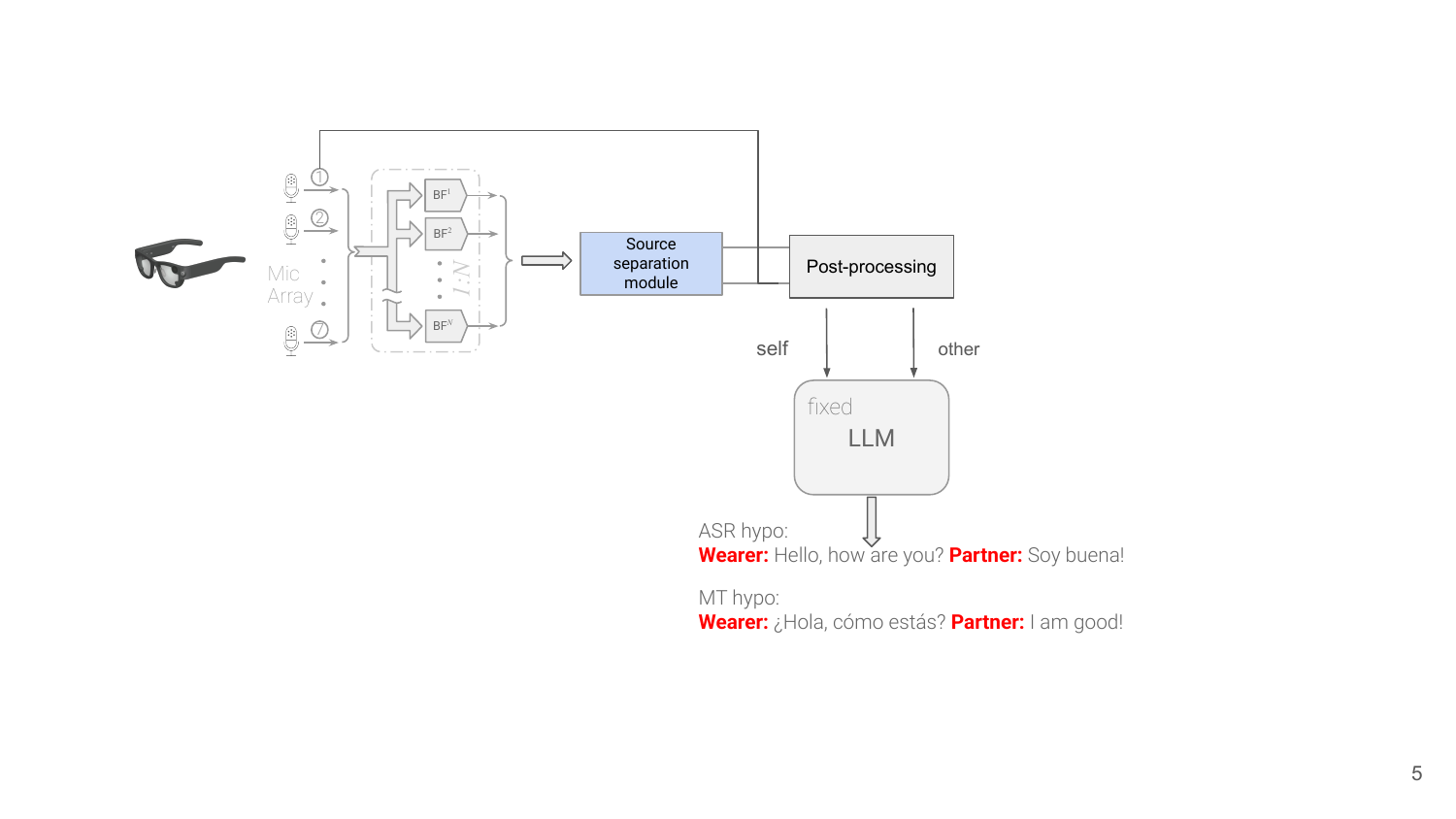}
    \caption{An illustration of a cascaded directional system comprising Source Separation and SLMs.}
    \label{fig:ss_gemma}
\end{figure}

As illustrated in Figure~\ref{fig:ss_gemma}, a two-stage system is proposed to enable SLMs to understand directional speech, comprising a source separation (SS) module and a post-processing module. Our source separation neural network is based on an encoder-decoder architecture~\cite{lin2022speech}. The process begins with the extraction of Short-Time Fourier Transform (STFT) features from the K + 1 beamformed channels. These time-frequency features are then fed into the encoder module, which consists of multiple convolutional blocks with gated linear units (GLU) activation~\cite{dauphin2017language} function and Dropout layers in between. The encoding output is then passed through a 3-layer LSTM, followed by a set of convolutional decoding layers. The decoder output is subsequently sent to a gating function, which returns the STFT masks associated with the wearer and partner speech from the reference audio. In our proposed architecture, the first audio channel is used directly as the reference audio. The final step involves computing the masked time-frequency outputs corresponding to the wearer and partner, which are then converted into the respective speech signals using the inverse STFT. The optimization objective for source separation modeling is a combination of three loss functions: L1 loss, STFT loss, and Log SI-SDR loss. This multi-objective approach enables the network to effectively separate the wearer and partner speech, resulting in improved speech quality and intelligibility.

However, source separation models inevitably introduce speech distortions for downstream tasks. To mitigate this, we refrain from using the separated audio directly as input to the SLMs. Instead, we utilize the separated audio to predict speaker tags on a chunk-by-chunk basis (e.g., 600ms chunk). Given that our focus is on conversations between wearer and partner speakers, we leverage a simple yet effective approach: comparing the Root Mean Square (RMS) ratio between the separated 2-channel signals, comprising wearer and partner channels, to determine which speaker is dominant in each chunk. Furthermore, a voice activity detection model is employed to filter out silent segments. The resulting speaker tag is then utilized to triage the appropriate prompt for various tasks. In summary, the final speaker tag is determined using the following equation:
\begin{equation}
\text{rms\_ratio} =  \frac{\mathrm{RMS}(\text{self})}{\mathrm{RMS}(\text{other})} > \alpha
\end{equation}
$\alpha$ is threshold to decide when to be self speaker. Once the speaker tag is assigned, the reference channel (a single-channel signal) is fed into the SLMs, which then process the corresponding prompts. However, this approach has a limitation in that it cannot handle overlapping speech. This issue can be addressed using end-to-end methods, such as an SOT-based solution, which will be discussed in detail in the next section.
\begin{figure}[!h]

    \centering
    \includegraphics[width=0.95\linewidth]{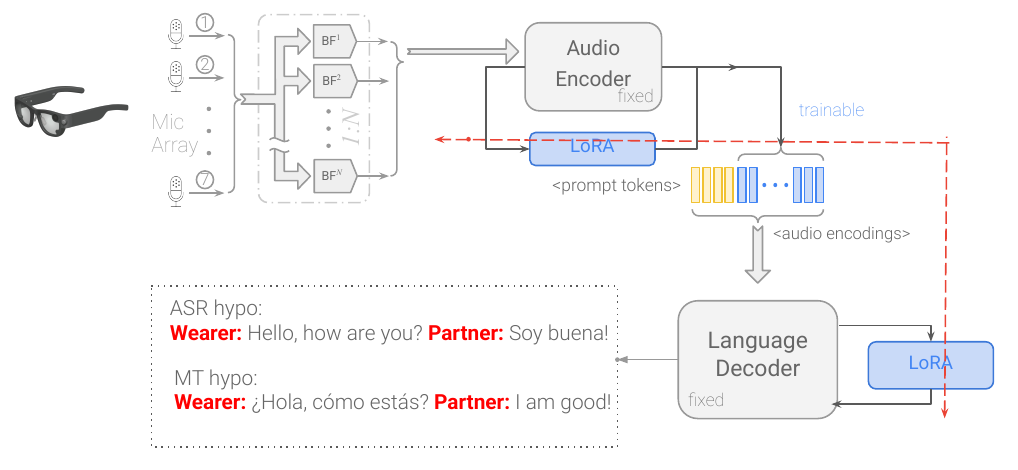}
    \caption{An illustration of MC-SOT based SLM post-training.}
    \label{fig:sot_gemma}
    \vspace{-5mm}
\end{figure}
\subsection{Serialized Output Training Post-Training}
Another way to enable SLMs to detect speaker changes is to incorporate SOT, a technique designed to identify speaker transitions between the wearer and a target speaker, as well as recognize partially overlapping speech. We fine-tune the SLMs with SOT targets using proper prompts, aiming to not only detect speaker changes but also perform specialized tasks, such as speech recognition or speech translation. 
To address directional speech recognition and speech translation, we have proposed a novel beamforming technique, as described in ~\cite{lin2024agadir}. Specifically, we utilize Non-Linearly Constrained Minimum Variance (NLCMV) beamforming to process multi-channel audio inputs. These beamformers employ predetermined coefficients and maintain a multi-channel structure, with each channel focusing on a distinct direction.

To preserve the original pre-trained audio encoder of the SLM, we select only one mouth beam as input for the SLM, rather than using multiple beams. The single mouth-beamformed signal already enhances the signal difference between wearer and partner signals. Previous studies have demonstrated the effectiveness of Low-Rank Adaptation (LoRA)~\cite{hu2022lora} in facilitating efficient fine-tuning of SLMs. As illustrated in Figure \ref{fig:sot_gemma}, we incorporate LoRA into our SOT post-training stage for both the audio encoder and decoder.

\section{Experiments and Results}
\subsection{Dataset}
Due to the lack of sufficient real-world multichannel training data, all multichannel training data in this work is simulated. The simulation is based on the geometry of a 5-microphone prototype modeled similar to the Aria glasses~\cite{engel2023project}. To introduce spatial diversity, we used room impulse responses (RIRs) sourced from real environments, simulating 12 distinct directions at 30° intervals. To better align with typical conversational scenarios, we defined five frontal directions of interest for partner: -60°, -30°, 0°, 30°, and 60°.
The simulated training data was created by spatially positioning single-channel audio clips to represent two primary roles: the wearer and their conversation partner. Additionally, we aimed to simulate bilingual conversations. For further details on the simulation process, refer to~\cite{lin2023directional}.
The training audio sources used for baseline systems and SLM finetuning are widely adopted datasets for speech processing: the Common Voice dataset\cite{ardila2019common} and the Multilingual LibriSpeech (MLS) dataset\cite{pratap2020mls}. The focus of this work is on four languages: English, French, Spanish, and Italian. For evaluation, the Fleurs dataset~\cite{conneau2023fleurs} was utilized. In order to train translation task, we applied a
teacher MT model to automatically translate the transcriptions
of common voice and multiligual librispeech into the target language. For training and testing the source separation model, the LibriSpeech dataset~\cite{panayotov2015librispeech} is utilized.
\vspace{-3mm}
\subsection{Model Setup}
To better understand the directional capabilities of SLMs, we employ two directional models as baselines. The first baseline is a multi-channel ASR system based on RNN-T~\cite{lin2023directional}, with a model configuration closely following the original work. For each beamformer direction, 80-dimensional log-Mel filterbank features are extracted and fed into a convolutional front-end. The front-end consists of two conv2d blocks with 5 channels, filters of size 2×5, and a stride setting of 1×2. The output is then stacked with six consecutive frames, forming a 512-dimensional vector, effectively reducing the sequence length by a factor of 6x. This is followed by 30 Emformer layers\cite{shi2021emformer}, each with 4 attention heads and 2048-dimensional feed-forward layers. The model has approximately 192M parameters.
The second baseline is JSTAR~\cite{moritz2025transcribing}, an end-to-end speech translation system that can output both ASR and machine translation (MT) in a streaming fashion. We use the same model architecture as detailed in~\cite{moritz2025transcribing}. The total number of JSTAR model parameters amounts to 121M, including the beamformer and feature extraction. Both directional baseline models are trained from scratch using the datasets mentioned above.

The SS model arctecture was adapted from ~\cite{feng2023directional}.  We extracted 257-dimensional complex SFTF for each beamformer direction or raw microphone
channel. Input features from multiple directions or channels are
concatenated. We use an Adam optimizer with a tri-stage learning rate scheduler. We trained the source separation models for 60 epochs, with a learning rate of 4e-4, a warmup of 10k iterations, and forced annealing after 10 epochs. The SS model size is about 6M parameters in total. During inference, the model operates in a chunk-wise decoding mode, with a chunk size of 600ms.

The SLM employed in this study is the open-source Gemma-3n 4B model~\footnote{https://huggingface.co/google/gemma-3n-E4B-it}, which is a instruction-tuned multimodal LLM that accepts multimodal prompts and responds in text. During the SOT fine-tuning stage with multi-channel simulated data, we set the LoRA rank to 64 for both the audio encoder and language decoder. This results in approximately 1.9\% of the total model parameters being updated.

\begin{table*}[!h]
\centering
\vspace{-6mm}
\begin{tabular}{lcccccccc}
\hline

\multicolumn{1}{l|}{\multirow{2}{*}{Model}}   & \multicolumn{4}{c|}{Wearer}                            & \multicolumn{4}{c}{Partner}        \\ \cline{2-9} 
\multicolumn{1}{l|}{}                         & WER[\%]$\downarrow$   & ins/del/sub & SA[\%]$\downarrow$  & \multicolumn{1}{c|}{BLEU$\uparrow$ } & WER[\%]$\downarrow$  & ins/del/sub  & SA[\%]$\downarrow$  & BLEU$\uparrow$  \\ \hline
\multicolumn{9}{c}{Wearer: English, Partner: Spanish}                                                                                       \\ \hline
\multicolumn{1}{l|}{Multi-channel ASR~\cite{lin2023directional}}        & 16.5 & 1.5/5.7/9.2 & 0.0 & \multicolumn{1}{c|}{-}    & 13.2 & 1.0/5.9/6.3  & 0.0 & -    \\
\multicolumn{1}{l|}{JSTAR~\cite{moritz2025transcribing}}                    & 16.7 & 1.3/5.9/9.4 & 0.0 & \multicolumn{1}{c|}{18.6} & 13.2 & 0.1/5.4/6.7  & 0.0 & 18.3 \\
\multicolumn{1}{l|}{SS+SLM}                 & 12.5 & 1.2/6.0/5.3 & 0.0 & \multicolumn{1}{c|}{22.0} & 10.6 & 1.0/5.0/4.6    & 0.6 & 25.3 \\
\multicolumn{1}{l|}{Multi-channel SOT+SLM} & 17.3 & 5.2/5.9/5.8 & 0.0 & \multicolumn{1}{c|}{19.6} & 16.5 & 1.7/7.3/5.7  & 1.8 & 22.6 \\ \hline
\multicolumn{9}{c}{Wearer: English, Partner: French}                                                                                        \\ \hline
\multicolumn{1}{l|}{Multi-channel ASR~\cite{lin2023directional}}        & 16.7 & 1.5/5.9/9.3 & 0.0 & \multicolumn{1}{c|}{-}    & 22.0 & 1.1/8.1/12.7 & 0.0   &   -   \\
\multicolumn{1}{l|}{JSTAR~\cite{moritz2025transcribing}}                    & 16.5 & 1.4/5.7/9.4 & 0.0 & \multicolumn{1}{c|}{19.3} & 21.9 & 1.1/7.7/13.2 & 0.0 & 16.5 \\
\multicolumn{1}{l|}{SS+SLM}                 & 12.9 & 1.0/6.7/5.2 & 0.0 & \multicolumn{1}{c|}{36.9} & 22.8 & 1.8/8.4/12.2 & 0.4 & 28.4 \\
\multicolumn{1}{l|}{Multi-channel SOT+SLM} & 15.5 & 2.8/7.1/5.5 & 0.2 & \multicolumn{1}{c|}{34.7} & 25.9 & 2.5/9.3/11.4 & 2.5 & 26.5 \\ \hline
\multicolumn{9}{c}{Wearer: English, Partner: Italian}                                                                                       \\ \hline
\multicolumn{1}{l|}{Multi-channel ASR~\cite{lin2023directional}}        & 16.1 & 1.4/5.8/8.9 & 0.0 & \multicolumn{1}{c|}{-}    & 22.3 & 0.8/14.6/6.8 & 0.0 & -    \\
\multicolumn{1}{l|}{JSTAR~\cite{moritz2025transcribing}}                    & 16.5 & 1.7/5.6/9.1 & 0.0 & \multicolumn{1}{c|}{19.3} & 17.6 & 1.0/7.9/8.6  & 0.0 & 16.5 \\
\multicolumn{1}{l|}{SS+SLM}                 & 13.3 & 1.6/6.4/5.3 & 0.0 & \multicolumn{1}{c|}{22.7} & 10.2 & 1.0/5.2/3.9  & 0.0 & 24.8 \\
\multicolumn{1}{l|}{Multi-channel SOT+SLM} & 16.8 & 4.3/6.3/5.7 & 0.0 & \multicolumn{1}{c|}{21.6} & 15.4 & 1.3/6.8/5.1  & 2.3 & 21.1 \\ \hline
\end{tabular}

\caption{ Speaker attribution (SA) error and attributed (”wearer”, ”partner”) word error rates (WER) on simulated FLEURS test data.}
\label{tab:gemma}
\vspace{-5mm}
\end{table*}
\vspace{-3mm}
\subsection{Evaluation Metrics}
The source separation model is evaluated using the commonly used \textem{perceptual
evaluation of speech quality} (PESQ) score~\cite{rix2001perceptual,ITU-P.862,ITU-P.863}, the \textem{short-time objective intelligibility} (STOI) score~\cite{taal2011algorithm} and scale-invariant SDR (SI-SDR)~\cite{le2019sdr}. The directional system used here also has the task of identifying who is speaking (Wearer or Conversation partner). This is reflected by a modified word-error rate (WER) metric, which, in addition to Substitution, Insertion, and Deletion errors, also counts words attributed to the wrong speaker as an error. This is denoted by SA for ``speaker attribution" error. For translation task, BLEU~\cite{papineni2002bleu} score is used to evaluate the model performance.
\vspace{-4mm}
\subsection{SLM Streaming Decoding}
Simultaneous speech translation and recognition using this pre-trained SLM is limited by two primary constraints: the models do not support true streaming input, requiring turn-based audio processing, and each inference is restricted to a maximum of 30 seconds of audio. To approximate a streaming experience without retraining the SLM, we segment the incoming audio stream into 600 ms chunks, which are optionally processed by a source separation model to distinguish between speakers (e.g., wearer and partner). These chunks are then accumulated into a sliding window, ensuring that the total audio context provided to the SLM does not exceed 30 seconds. In parallel, we maintain a running history of the previously generated text, truncating this context to the most recent 50 words to provide relevant linguistic cues for the model. Although this method does not guarantee perfect alignment between the audio and text contexts, empirical results show that the SLM can still produce coherent and accurate transcriptions and translations. For each inference, we issue two distinct prompts—one instructing the model to transcribe the audio in the source language and another to translate the audio into the target language—enabling the system to deliver both ASR and translation outputs concurrently with low latency and high responsiveness.
\begin{table}[]
\centering
\setlength{\tabcolsep}{4pt}
\resizebox{0.65\columnwidth}{!}{
\begin{tabular}{l|cc|cc}
\hline
\multirow{2}{*}{Metrics} & \multicolumn{2}{c|}{Wearer}   & \multicolumn{2}{c}{Partner}   \\ \cline{2-5} 
                         & Mixed & Separated  & Mixed & Separated \\ \hline
PESQ$\uparrow$                     & 1.60        & 2.91            & 1.52        & 1.74            \\ \hline
STOI$\uparrow$                     & 0.91        & 0.97            & 0.70        & 0.81            \\ \hline
SI-SDR$\uparrow$                   & 5.85        & 19.56           & -13.28      & 8.66            \\ \hline
\end{tabular}
}
\caption{Source Separation model performance evaluation in terms of PESQ, STOI and SI-SDR.}
\label{tab:ss_perf}
\vspace{-7mm}
\end{table}

\subsection{Results}
\vspace{-2mm}
We first evaluated the SS model's performance. As shown in Table~\ref{tab:ss_perf}, we observed that the SS model performs well on both wearer and partner speaker separation in terms of all three metrics used in this study. For example, the SI-SDR of the partner speaker is significantly improved from -13.28 dB to 8.66 dB. This indicates that the SS model is highly effective in separating and enhancing the target speaker's audio, while suppressing background noises.

Next, we conducted a comparative analysis of the performance between baseline and proposed methods, focusing on speech recognition and speech translation. The results, presented in Table~\ref{tab:gemma}, indicate that SLM-based directional systems exhibit superior directional understanding capabilities in terms of speaker attribution errors. In comparison to directional baseline systems tailored for smart glasses, the SA of both SS+SLM and SOT+SLM on the simulated multi-channel FLEURS test set is nearly perfect for wearer, matching the performance of the baseline systems. However, we observed a higher rate of attribution errors for Partner when using SOT-based SLM. Nevertheless, the error rate remains below 2.5\% for all language pairs. This was primarily due to the issue that SLM does not follow instructions well, likely because we fine-tuned the model to output both transcriptions and translations simultaneously. The model appears to be confused between transcribing and translating, which may be causing it to struggle with accurate attribution. In terms of BLEU score, both proposed systems outperform both the multi-channel ASR and JSTAR systems. Specifically, on the "Partner speaking Spanish" scenario, SS+SLM and SOT+SLM improve the BLEU score from 18.3 to 25.3 and 22.6, respectively, compared to JSTAR.

For the speech recognition task, the SS+SLM solution appears to perform the best among all the systems. However, in some cases, the SOT+SLM solution exhibited underperformance compared to the baseline systems. A detailed error analysis revealed that the degradation in performance was primarily caused by speaker attribution errors. The speaker attribution errors led to a higher rate of deletion errors for 'Partner' and also resulted in a higher rate of insertion errors for 'Wearer'. To further improve performance, one potential approach is to explore alternative strategies that enable the model to accurately follow instructions.

A comparison of the two proposed systems reveals that SS+SLM outperforms SOT+SLM in both speech recognition and speech translation tasks. However, it is important to note that the SS+SLM approach has a limitation in its current form, as it is unable to handle overlapping speech. This approach is well-suited for speech translation use cases, as one can reasonably expect limited overlap in conversations—participants typically wait for the translation before responding to others. In contrast, one of the key advantages of SOT+SLM is its ability to effectively handle overlapping speech from multiple talkers.
\vspace{-6mm}
\section{Conclusion}
\vspace{-3mm}
In this paper, we presented two approaches to endow SLM with directional understanding capabilities tailored for smart glasses scenarios. Our experimental results demonstrate that the SLM-based solution shows an ability to accurately distinguish between wearer (near-field) and partner (far-field) speech in terms of speaker attribution error. Additionally, the proposed source separation-based SLM and SOT-based SLM exhibit strong performance in both speech recognition and speech translation tasks.

{\footnotesize
\bibliographystyle{IEEEbib}
\bibliography{strings,refs}}

\end{document}